%% file: main.tex
\documentclass[10pt,twocolumn,letterpaper]{article}

\usepackage[pdftex]{graphicx}
\pdfoutput=1
\usepackage{cvpr}              % To produce the CAMERA-READY version
\input{preamble}

\definecolor{cvprblue}{rgb}{0.21,0.49,0.74}
\usepackage[pagebackref,breaklinks,colorlinks,allcolors=cvprblue]{hyperref}

\def\paperID{7384} % *** Enter the Paper ID here
\def\confName{ICCV}
\def\confYear{2025}

\title{Embodied Navigation with Auxiliary Task of Action Description Prediction} 

\author{
Haru Kondoh$^{1}$ \quad
Asako Kanezaki$^{1,2}$\\[0.3em]
$^{1}$ Institute of Science Tokyo \quad
$^{2}$ RIKEN AIP\\[0.3em]
{\tt\small kondo.h.4aa3@m.isct.ac.jp} \quad
{\tt\small kanezaki@comp.isct.ac.jp}
}

\begin{document}
\maketitle

\input{sec/0_abstract}    
\input{sec/1_intro}
\input{sec/2_related}
% \input{sec/3_task}
\input{sec/4_method}
\input{sec/5_experiment}
\input{sec/6_conclusion}

%\section*{Acknowledgments}
\noindent \textbf{Acknowledgments: }
This work was supported by JST FOREST Program, Grant Number JPMJFR206H.

{
    \small
    \bibliographystyle{ieeenat_fullname}
    \bibliography{main}
}


% {
%     \small
%     \bibliographystyle{ieeenat_fullname}
%     \bibliography{main}
% }

\end{document}

%% file: preamble.tex
\usepackage{lineno}

%% file: sec/0_abstract.tex
\begin{abstract}
The field of multimodal robot navigation in indoor environments has garnered significant attention in recent years. However, as tasks and methods become more advanced, the action decision systems tend to become more complex and operate as black-boxes. 
For a reliable system, the ability to explain or describe its decisions is crucial; however, there tends to be a trade-off in that explainable systems cannot outperform non-explainable systems in terms of performance.
In this paper, we propose incorporating the task of describing actions in language into the reinforcement learning of navigation as an auxiliary task.
%In addition,
% 模倣学習の文脈では自分に与えられている指示文を予測することについては取り組まれているが、
%it has been difficult to let reinforcement learning (RL) models to generate action descriptions due to the lack of ground-truth data. We solve this problem by successfully knowledge distillation from foundation models, etc., into a navigation RL model. 
Existing studies have found it difficult to incorporate describing actions into reinforcement learning due to the absence of ground-truth data.
We address this issue by leveraging knowledge distillation from pre-trained description generation models, such as vision-language models.
%In this paper, we defined several action descriptions and conducted a comprehensive investigation using several semantic navigation tasks, including the particularly challenging multimodal navigation called semantic audio-visual navigation. Ultimately, our method is descriptive and demonstrated state-of-the-art performance.
We comprehensively evaluate our approach across various navigation tasks, demonstrating that it can describe actions while attaining high navigation performance. Furthermore, it achieves state-of-the-art performance in the particularly challenging multimodal navigation task of semantic audio-visual navigation.

% The explainability of complex systems is important; however, it is known that there often exists a trade-off where explainable systems tend to perform less effectively compared to inexplicable systems. In this paper, we propose incorporating the task of explaining robot actions through language into the reinforcement learning of navigation as an auxiliary task. We show that the task of verbalizing action predictions is in fact closely related to reinforcement learning of navigation, and that learning these tasks in a complementary manner can also improve navigation performance. In several semantic navigation tasks, including the particularly challenging multimodal navigation called semantic audio-visual navigation, our method improved upon multiple baseline approaches, ultimately demonstrating state-of-the-art performance.

\end{abstract}

%% file: sec/1_intro.tex
\section{Introduction}
\label{sec:intro}

\begin{figure}[t]
    \begin{center}
        \centering
        \includegraphics[width=\linewidth]{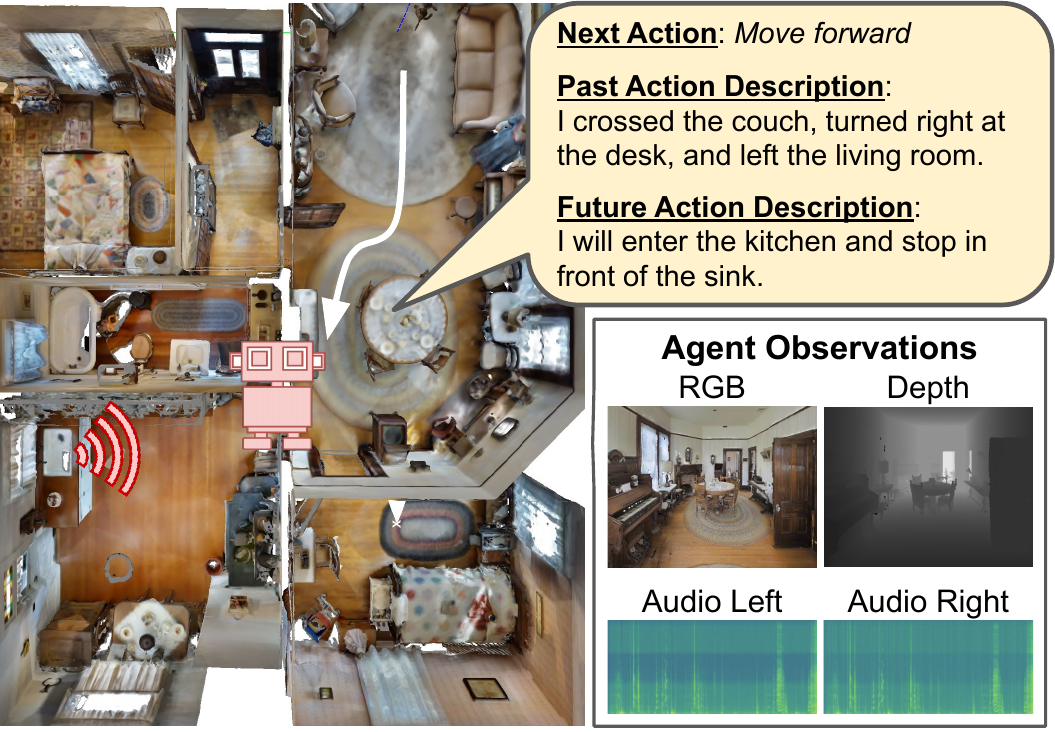}
        \caption{
            Overview of this study.
            %%The robot must navigate to a sounding object based on the observations of visual and auditory information.
            %%The robot must also generate an explanation of what it should do in the future or what it has done in the past.
            The robot generates action descriptions of its past actions or future plans during navigation, such as moving toward a sounding object in the case of semantic audio-visual navigation, based on visual and auditory observations.
        }
        \vspace{-5mm}
        \label{fig:reliable_semantic_audio_visual_navigation}
    \end{center}
\end{figure}

The development of robots that recognize their environment and autonomously navigate to a specified target has received particular attention in the last decade.
%Most of them are navigation tasks in which the robot can only observe visual information.
Recently, multimodal navigation tasks have been proposed that can handle language \cite{anderson2018vision,qi2020reverie,zhu2021soon} and can observe auditory information \cite{gan2020look,chen2020soundspaces}.
Furthermore, various methods have been proposed to improve navigation performance, such as end-to-end reinforcement learning (RL) 
 methods \cite{fang2019scene,anderson2018vision,chen2020soundspaces} and methods that utilize large language models (LLMs) \cite{majumdar2022zson,zhou2024navgpt,yang2024rila}.
%As multimodalization and performance improvements advance, 
As tasks become more complex and challenging,
models become increasingly complex and black-boxed, making explainability crucial to addressing resulting lack of transparency.
% However, there are very few studies dealing with explainability in navigation. In addition,
However, in general, prior research on explainability has been plagued by the trade-off between explainability and the performance of the main task \cite{freitas2014comprehensible,8400040,rudin2019}.
Using imitation learning (IL), there have been attempts to predict the instruction sentences given to the model \cite{zhu2020vision,hejna2023improving}, in the context of instruction following tasks such as vision-and-language navigation \cite{anderson2018vision}. However, the lack of ground-truth sentences in RL makes it difficult to extend these IL methods to RL, and this fact has limited the range of applications. 

We propose a method to describe the system's actions in natural language, aiming for an explainable navigation system. We define multiple types of action descriptions and focus on the affinity between this action description generation learning and navigation learning.
% Previously, affinities between different visual tasks, such as edge detection and depth estimation, have been comprehensively investigated \cite{zamir2018taskonomy,standley2020tasks}, revealing the existence of effective task combinations.
% We also found that a similar affinity exists between navigation and action descriptions.
Surprisingly, by formulating the action description as an auxiliary task through knowledge distillation from pre-trained models, such as large vision-language models (VLMs), to RL models, we developed a method that enables RL agents to describe actions without compromising navigation performance. % Furthermore, we suggest that action descriptions may become an element of explainability of the system in the future by making it possible to analyze the causes of failure.

% Furthermore, we propose a method of knowledge distillation from a foundation model to the navigation agent by treating the output of the foundation model as teacher data for the explanation generation task. This method allows the navigation agent to implicitly acquire knowledge that is important for navigation, without the need for high-cost human-generated data.

Figure \ref{fig:reliable_semantic_audio_visual_navigation} represents an overview of this research.
% We address action descriptions in several navigation tasks including Semantic Audio-visual Navigation (SAVNav) \cite{chen2021semantic}, a particularly challenging task in multimodal navigation.
% The robot must navigate to a given goal by observing first-person visual and auditory information.
% Furthermore, the robot must describe what it is about to do in the future or what it has done in the past in natural language.
% This task has a variety of applications, ranging from commercial and domestic use to lifesaving.
% The more descriptive the robot is, the greater the trust users will have in it, leading to broader real-world applications.
% Also, this will facilitate smoother collaboration with humans.
% Furthermore, as shown in section \ref{sec:experiments}, it also has the advantage that it is possible to analyze why the navigation agent made the mistakes and to indicate directions for improvement.
%
During navigation, the RL agent also needs to describe what it has done in the past or what it should do in the future while making action decisions based on observations.
We demonstrate that the proposed method is applicable not only to a specific task (e.g., instruction following) but also to a wide range of navigation tasks, and it achieves state-of-the-art performance in a particularly challenging task, semantic audio-visual navigation (SAVNav) \cite{chen2021semantic}.
We also show that the proposed method makes the RL model descriptive and allows us to analyze failure cases.
The main contributions of this paper are summarized as follows.
\begin{itemize}

\item Focusing on the compatibility between action description and navigation, we propose an RL method to learn both descriptive and high-performing navigation models.
\item By leveraging pre-trained VLMs, our method removes the reliance on human-created data. 
%%This approach broadens its applicability.
\item A comprehensive experiment of various action descriptions and auxiliary task methods is conducted to analyze the proposed method.
%\item Comprehensive experiments using Soundspaces \cite{chen2020soundspaces} showed that the process of learning explanation generation contributes to learning navigation. 
%%We also show the current limitations in the explainability of the proposed method.
\item Through comparisons across various navigation tasks and existing methods, we demonstrate the consistent effectiveness of the proposed approach, achieving performance surpassing the state-of-the-art in the particularly challenging SAVNav task.
\end{itemize}

%% file: sec/2_related.tex
\section{Related Work}
\label{sec:related_work}
%\vspace{-2mm}
\noindent \textbf{Explainable Reinforcement Learning (XRL)}
is the framework where a reinforcement learning agent communicates its situation and reasons the decisions in a way humans can understand.
Traditional approaches often involve transforming tasks into hierarchical structures by dividing them into smaller and more understandable subtasks for humans \cite{shu2018hierarchical,10.1109/IROS40897.2019.8968488}, representing networks as tree structures \cite{Coppens2019DistillingDR,roth2019conservative,liu2019toward}, visualizing where the agent focuses on \cite{juozapaitis2019explainable,HE2021107052,skirzynski2021automatic,mott2019towards,Leurent2019SocialAF}, or generating explanations in natural language \cite{ehsan2018rationalization,ehsan2019automated}. Recently, methods utilizing large language models (LLMs) to generate explanations related to agents' actions and observations have also been proposed \cite{10.1007/978-3-031-40725-3_45,wang2023voyager,yao2023react}. Also, Stein \etal \cite{stein2021generating} addressed explainability in navigation by generating explanatory sentences in a rule-based method in the form of fill-in-the-blanks.
%
%%In this study, we propose an approach for generating explanations in natural language. In prior work, the explanation generation module was often entirely decoupled from the policy, meaning that the generated explanations did not necessarily reflect the agent's decision-making process. Moreover, many c
Conventional methods have focused solely on explainability or interpretability, leading to a trade-off with the performance of the main task \cite{freitas2014comprehensible,8400040,rudin2019}. Similarly to previous studies~\cite{ehsan2018rationalization,ehsan2019automated}, we adopt a method that describes action decisions in natural language. However, our approach differs from theirs in that it integrates the action description generation module with the policy, which not only enhances the quality of the generated action description but also improves the performance of the navigation task.

\noindent \textbf{Vision and Language Navigation (VLN)}
is a task of navigation based on first-person visual observations to follow instructions in a given natural language \cite{anderson2018vision}.
Previous research 
%has extensively explored methods to tackle this task. These 
include splitting long instructions into shorter segments \cite{he2023mlanet,zhu2020babywalk}, effectively integrating historical and multimodal information \cite{krantz2020beyond,chen2021history,hong2021vln}, optimizing the use of the topological map \cite{chen2022think,an2023bevbert,an2024etpnav}, and enhancing performance by augmenting trajectory-instruction paired data \cite{tan-etal-2019-learning,fried2018speaker,Wang2021LessIM,kamath2023new,wang2023scaling}. Recently, methods for fine-tuning large pre-trained models \cite{lin2024navcot,pan2024langnav,zheng2024towards} or utilizing them in a zero-shot manner \cite{shah2023lm,zhou2024navgpt,Zhou2024NavGPT2UN} have also been proposed. The above approaches primarily use instructions as only inputs. Similar to our work, Zhu \etal \cite{zhu2020vision} and Hejna \etal \cite{hejna2023improving} use given instructions as outputs as well, treating instruction prediction as an auxiliary task. The most significant difference from our work is that they use only imitation learning, which has limitations such as a restricted exploration space, use of expensive demonstration data and dependency on human demonstration capabilities.
It is not possible to simply transfer their method to RL because the ground-truth pair data for trajectory and instruction is not available in RL.
%%This study proposes a method to challenge and solve this problem.
%%Moreover, b
%%By applying knowledge distillation \cite{hinton2015distilling} from large vision language models, we enable instruction prediction as an auxiliary task without human-generated instruction data. Therefore, our approach can be applied to any reinforcement learning and imitation learning environment that involves image-based tasks.

\noindent \textbf{Audio Visual Navigation}
is a task that involves navigation to a sound location by observing auditory information in addition to visual information.
Various RL-based methods have been proposed to tackle this task~\cite{chen2020soundspaces,chen2021semantic,majumdersemantic,chen2020learning,Yu_2022_BMVC,dean2020see,chen2023omnidirectional,wang2023learning,yu2022sound,tatiya2022knowledge}.
% Various approaches have been proposed, including methods for learning navigation policies through reinforcement learning \cite{chen2020soundspaces,chen2021semantic,majumdersemantic}, effective use of maps and waypoints generation \cite{chen2020learning,chen2023omnidirectional}, integration of visual and auditory multimodal information \cite{Yu_2022_BMVC}, and the use of intrinsic rewards to promote exploration \cite{dean2020see}. Additional strategies include shifting point goal navigation methods to improve sample efficiency \cite{chen2023omnidirectional}, focusing solely on the spatial information of sound \cite{wang2023learning}, applying adversarial learning of a moving goal \cite{yu2022sound}, and leveraging prior knowledge through knowledge graphs \cite{tatiya2022knowledge}.
While these methods do not incorporate language, Paul \etal \cite{paul2022avlen} and Liu \etal \cite{Liu2023dec2} introduced language into audio-visual navigation as instructions to reach the goal.
%Paul \etal \cite{paul2022avlen} were the first to introduce language into audio-visual navigation. 
%%Their approach used language only as an input for the robot. However, 
%Liu \etal \cite{Liu2023dec2} later proposed a method that utilizes language not only as an input but also as an output. 
More recently, methods leveraging LLMs have been proposed to more effectively utilize language in solving this task \cite{huang2023audio,yang2024rila}. 
%%These methods have demonstrated remarkable navigation performance. 
Notably, Yang \etal \cite{yang2024rila} achieved higher success rates compared to learning-based methods. However, in terms of the SPL metric, which is considered crucial for navigation evaluation \cite{anderson2018evaluation}, their performance fell short of that of simple learning-based baselines.

% In this study, we propose a learning-based approach that uses natural language only as an output. The significant difference between our method and previous language-based approaches is that we do not separate the language generation module from the policy module. We focus on the synergistic effect of integrating these two modules, ultimately significantly improving navigation performance.

%% file: sec/4_method.tex
\section{Proposed Method}
\label{sec:method}
%\vspace{-2mm}

%%\subsection{Overview}

\begin{figure*}[t]
    \begin{center}
        \centering
        \includegraphics[width=\linewidth]{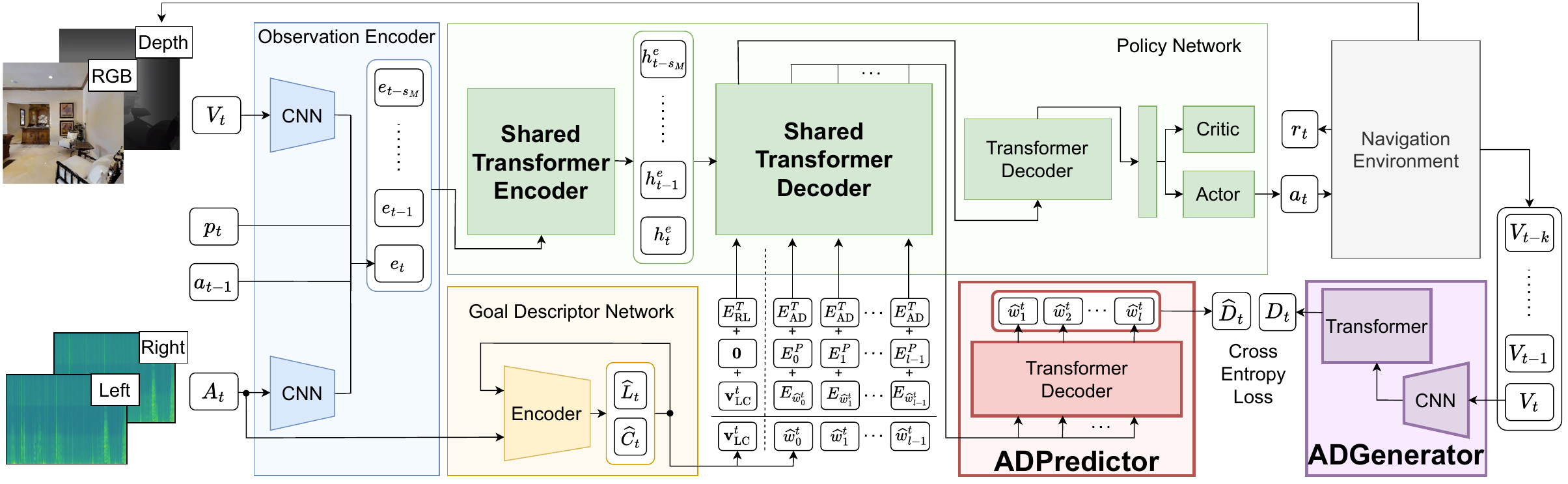}
        \vspace{-2em}
        \caption{
            Overview of DescRL applied to SAVi \cite{chen2020soundspaces}, which is a method for semantic audio-visual navigation.
            The RL agent receives visual observation $V_t$, auditory observation $A_t$, posture $p_t$, and previous action $a_{t-1}$ at time $t$ and outputs the next action $a_t$ and the action description $\hat{D}_t$. $E^T$ and $E^P$ represent task embedding and positional encoding, respectively.
            Here, ADPredictor is trained as an auxiliary task to predict the output $D_t$ of the pre-trained ADGenerator.
        }
        \vspace{-4mm}
        \label{fig:eprl_arhitecture}
    \end{center}
\end{figure*}

We propose descriptive reinforcement learning (DescRL), a method that enables to describe its own actions and improves navigation performance by treating action description as an auxiliary task.
%%Most conventional navigation methods learn to output only the next action based on previous observations.
%%Our method does not only learn about the next action but also learns to explain what it has done in the past or what it should do in the future.
% Our method is inspired by Mirowski \etal \cite{mirowski2016learning}, which also improves navigation performance by allowing it to predict depth images from a given RGB image.
%
% The enhancement of navigation performance through learning to generate explanations can be attributed to two primary reasons: the acquisition of common sense and the enhancement of planning abilities.
% For example, suppose there is a sound coming from a sink in an indoor environment (Fig. \ref{fig:reliable_semantic_audio_visual_navigation}).
% The robot says ``Sounds coming from the sink. The sink is likely to be in the kitchen. So I head for the kitchen first.''
% In this case, the agent can learn about common sense, such as, ``The sink is likely to be in the kitchen.'' The agent can also learn about planning to explain ``So I head for the kitchen first.''
% In other words, the acquisition of common sense and the improvement of planning ability, which are important elements for navigation, can be realized by learning explanation generation.
%
Here, we deal with three types of action description, past action description (P-AD), future action description (F-AD) and past-future action description (PF-AD) (Fig.~\ref{fig:reliable_semantic_audio_visual_navigation}). P-AD provides a verbalization of what the agent has done in the past. This helps to identify recognition errors in objects or spaces, or to verify that past information important for navigation is accurately remembered. Learning to predict P-AD will contribute to recognizing objects, spaces, etc., attention to important information, and acquiring common knowledge such as ``the couch and the desk are in the living room.'' F-AD provides a verbalization of what the agent should do in the future. This helps to verify whether the agent knows exactly what actions to take in the future, or whether the agent knows exactly but is unable to link them to actual actions. Learning to predict F-AD will contribute to the improvement of planning abilities and the acquisition of common sense, such as ``the sink is in the kitchen.'' Also, PF-AD is a combination of P-AD and F-AD. This provides a verbalization what the agent has done in the past and what the agent should do in the future.

DescRL is especially important for a complex task such as Semantic Audio-visual Navigation (SAVNav) \cite{chen2021semantic}.
In this task, the goal location must be identified even when auditory information is not observable because the sound may stop in the middle of the episode.
% Common sense about where the goal object will likely be can mitigate this problem.
In addition, if the sound stops in the middle of the episode, RL for navigation becomes difficult because the current observation is no longer enough to provide clues about the goal and cannot infer the reward.
From this point of view, it is important to use auxiliary tasks for continuous learning. Experimentally, we show that DescRL is a particularly good auxiliary task compared to other auxiliary tasks.

Our proposed method is divided into two major phases.
In phase 1, the ADGenerator, a module that verbalize navigation observations (bottom right of Fig. \ref{fig:eprl_arhitecture}), is pre-trained to generate action descriptions from navigation observations.
In phase 2, the navigation is trained, while the action description prediction is treated as an auxiliary task.

\vspace{-1.5mm}
\subsection{Phase 1: Pre-train ADGenerator}
\vspace{-1.5mm}

We first pre-train the ADGenerator, which takes the sequence of visual information observed by the agent and translates it into language.
This is similar to a video captioning model or the speaker-model \cite{fried2018speaker}.

\noindent \textbf{Structure in Object-goal Navigation (ObjNav) and SAVNav:}
% The structure of the XGenerator used in this study is described below.
The ADGenerator receives a one-hot vector sequence of the agent's actions $\bold{a}_1, \dots, \bold{a}_T \in \{0, 1\}^4$ and the observed visual information sequence $V_0, \dots, V_T \in \mathbb{R}^{128\times 128 \times 7}$.
Here, in addition to RGBD images that can be observed by the RL agent, the ADGenerator also uses semantic images. The ADGenerator passes the visual information sequence through a convolutional neural network $f_V^g : \mathbb{R}^{128\times 128 \times 7} \rightarrow \mathbb{R}^{512}$ to obtain the visual feature sequence $f_V^g (V_0), \dots, f _V^g(V_{T-1})$.
Then, this visual feature sequence and the action sequence are concatenated to obtain $(f_V^g(V_0), \bold{a}_1), \dots, (f_V^g (V_{T-1}), \bold{a}_T)$.
Finally, by inputting this concatenated feature sequence to the transformer \cite{vaswani2017attention} encoder, the word sequence $w_1, \dots, w_{l}$ is output from the transformer decoder.
This word sequence is the verbalization of the visual information sequence $V_0, \dots, V_T$.

\noindent \textbf{Structure in VLN:}
In VLN, the agent navigates on a graph. The ADGenerator receives the observation information $v_0, \dots, v_K$ for each observed node and their neighbor information $\epsilon_t$, which indicates whether each pair of nodes is adjacent. These are input to an encoder, which has the same structure as the Coarse-scale Cross-modal Encoder except for the Cross-Attention in DUET \cite{chen2022think}. Then, the visual information features $\hat{v}_0, \dots, \hat{v}_ K$ are obtained. Finally, by inputting these information into the transformer decoder, the sentence $w_1, \dots, w_{l}$ is obtained.

\noindent \textbf{Training details: }
The ADGenerator was trained using the R2R dataset \cite{anderson2018vision,krantz2020beyond}, which is a human-created dataset for vision and language navigation.
The $i$-th data contains the navigation language instructions $w_1^i, \dots, w^i_{l_i}$, a sequence of visual information $V_0^i, \dots, V_{T_i}^i$ and an action sequence $\bold{a}_1^{i}, \dots, \bold{a}_{T_i}^{i}$ that can be observed by navigating along the given language instruction.
So, the ADGenerator is trained to verbalize how the agent navigated from the input observations.
We used 10,819 pieces of data for training and 1,839 pieces of data for evaluation.
The training was performed using the teacher-forcing method \cite{10.1162/neco.1989.1.2.270}.
The cross entropy loss was used as the loss function.
As the word embedding, GloVe~\cite{pennington2014glove} was used for ObjNav and SAVNav, and BERT \cite{Devlin2019BERTPO} tokenizer was used for VLN.
See the supplementary materials for qualitative evaluation.

\vspace{-1.5mm}
\subsection{Phase 2: Action Description Prediction for Reinforcement Learning}
\label{subsec:eprl}
\vspace{-1.5mm}

After pre-training the ADGenerator, the RL model learns navigation while treating action description prediction as an auxiliary task.
The module 
%that predicts explanations in the reinforcement learning model 
is called ADPredictor.
The RL model learns to output appropriate actions from the policy network as before but also learns to output action descriptions from the ADPredictor.
Figure \ref{fig:eprl_arhitecture} shows the overview of the application of DescRL to SAVi \cite{chen2021semantic}, which is a method for SAVNav.
SAVi has a CNN-based observation encoder, a transformer-based policy network, and a goal descriptor network that predicts goal location $L_g$ and category $C_g$.
Concatenating them, the input to the policy network $v_\mathrm{LC}^t = (\hat{L}_t, \hat{C}_t)$ is calculated.
% In this study, XPredictor and XGenerator are added to this.
ADPredictor consists of a transformer decoder.
By inputting the predicted goal location $\hat{L}_g$ and category $\hat{C}_g$ of the goal as the beginning of sentence ($\hat{w}_0^t$ in Fig. \ref{fig:eprl_arhitecture}), an action description is predicted considering $\hat{L}_g, \hat{C}_g$.
Here, the transformer encoder and decoder are shared by ADPredictor and the policy.
This allows the shared encoder and decoder to be trained on two tasks so that it can encode observation information more effectively. Also, it is expected to be able to predict more faithful \cite{lyu_etal_2024_towards} action descriptions for the model since they share most of the weights.
% We have introduced task embeddings $E^T$ to share the decoder.
% This acts like the positional encoding commonly used in transformer. $E^T_\mathrm{RL}$ and $E^T_\mathrm{AD}$ are applied to the input for policy and ADGenerator, respectively.
To enable decoder sharing, task embeddings $E^T_\mathrm{RL}, E^T_\mathrm{AD}$ are applied to the input for policy and ADGenerator, respectively. This acts like the positional encoding commonly used in transformer.
In addition, visual observations for the past $k+1$ steps are input to the pre-trained ADGenerator.
As a result, the ADGenerator outputs an action description of what the agent has done in the past.
ADPredictor is trained with cross-entropy loss $\mathcal{L}^{\rm{CE}}$ to be able to predict this ADGenerator output.
Therefore, if the loss function for RL is $\mathcal{L}^{\rm{RL}}$, this model is trained to minimize $\mathcal{L}^{\rm{RL}} + \lambda \mathcal{L}^{\rm{CE}}$,
where $\lambda$ is a coefficient indicating how much to account for the loss of action description prediction.
For other architectures of other tasks, the ADPredictor and ADGenerator were applied in the same way (see the supplementary materials for more details). 
%%Please refer to the supplementary materials (section \ref{supp:network_architecture_details}) for more details.

The training of this RL model is divided into two steps.

\noindent \textbf{Step 1:} Pre-training of ADPredictor. Here, this model does not learn navigation but only action description prediction to learn the observation encoder, the shared encoder and decoder, and ADPredictor.

\noindent \textbf{Step 2:} By learning navigation and action description prediction at the same time, both the entire policy network and ADPredictor are learned.
Here, the ADPredictor is always trained in a teacher-forcing manner \cite{10.1162/neco.1989.1.2.270}.

For training in Step 1, R2R can be utilized in VLN, but not in SAVNav and ObjNav. Therefore, we created a new dataset with the following operations.
First, the observation $O_0, O_1, \dots, O_T$ is obtained by taking the shortest path from a random start to a random goal, and action sequences $\bold{a}_1, \dots, \bold{a}_T$ are collected.
Then, these information are input to the ADGenerator learned in phase 1 to obtain the ground truth action description $D$.
ADPredictor is trained using the dataset consisting of approx. 100k data for ObjNav and 500k data for SAVNav created in advance using this method.
The ADGenerator is used only during training and not during testing so that future observations can be input.
That is, instead of $V_{t-k}, \dots, V_t$, the visual observations $V_t, \dots, V_{t+k}$ obtained by following the shortest path to the goal can be input to ADGenerator.
Here, the output $D_t$ of the ADGenerator represents what to do in the future.
In this case, the ADPredictor must predict what to do in the future based on past observations and the predicted location and category of the goal.
In the following, we refer to DescRL that allows past/future/past-and-future observations to be input to the ADGenerator as Past-DescRL/Future-DescRL/Past-and-Future-DescRL (P-DescRL/F-DescRL/PF-DescRL), respectively.
Future observations are only used at train time through ADGenerator. At test-time, the ADPredictor only has access to past observations, ensuring fairness with prior studies
% Also, future observations are used only in the ADGenerator, and the ADGenerator is not used during testing because the ADPredictor is used during testing.
% This ensures that the RL model does not receive future observations at any point, maintaining fairness with respect to prior studies.
% Also, since ADPredictor is used to predict action descriptions during testing, future observations are not required during testing. 

\noindent \textbf{Training details in ObjNav: }
We used 8 GPUs and assigned 10 processes to each GPU, for a total of 80 processes for training. 
The RL algorithm used for training was DD-PPO \cite{wijmans2019dd}.
The number of parameter updates was set to 3,000.
The number of transformer encoder layers was set to 2, and the number of transformer shared decoder layers was set to 2. The number of unshared decoder layers was set to 1 in both the policy and ADPredictor.
The coefficient of DescRL loss was set to $\lambda = 0.1$, and the input length to ADGenerator was set to $k+1 = 20$.

\noindent \textbf{Training details in VLN: }
We used 1 GPU and assigned 1 process to the GPU for training. The imitation learning 
 (IL) algorithm DAgger \cite{ross2011reduction} was used for training. 
Our method is applicable not only to RL but also to IL.
The number of transformer shared decoder layers was set to 0 because baseline methods did not have any decoder. The number of unshared decoder layers was set to 3 in ADPredictor.
The coefficient of DescRL loss was set to $\lambda = 0.1$, and all visual observation history is input to ADGenerator.

\noindent \textbf{Training details in SAVNav: }
We used 4 GPUs and assigned 8 processes to each GPU, for a total of 36 processes for training. Other settings related to DescRL are the same as for ObjNav.
More details can be found in the supplementary materials
%(section \ref{supp:network_architecture_details})
and the code to be published.

\vspace{-1.5mm}
\subsection{Vision language models for ADGenerator}
\vspace{-1.5mm}

In the above, we proposed that the ADGenerator is trained from scratch on the human-created VLN dataset R2R \cite{anderson2018vision,krantz2020beyond}. This means that the proposed method still relies on human-created data, which undermines its applicability. Here, we also propose an alternative approach based on knowledge distillation techniques \cite{hinton2015distilling} by focusing on vision language models (VLMs) that have already acquired common sense from large-scale web data. 

We used VideoLLaMA2 \cite{damonlpsg2024videollama2} and QWen2.5-VL \cite{bai2025qwen2} as VLM. 
% Given a visual observation sequence of length $l^v$,
It receives an input consisting of RGB images $(V_1, \dots, V_{l^v}) \in \mathbb{R}^{l^v \times 336\times336\times3}$ and a prompt (see the suppulementary materials for details) The output is treated as the ground-truth data for ADPredictor's output. Knowledge distillation was conducted in the form of soft targets \cite{hinton2015distilling}. 
% Therefore, denoting the output of VideoLLaMA2 and XPredictor as $p^T, p^S$, respectively, the loss function for updating XPredictor can be expressed as $\mathcal{L}^\mathrm{CE} = - \frac{1}{L} \sum_{l=1}^L \sum_{w \in \mathrm{Vocab}} p^T(w_l=w|w_1,\dots,w_{l-1}) \log p^S(w_l=w|w_1,\dots,w_{l-1})$. Here, $L$ represents the length of the sentence, and $\mathrm{Vocab}$ denotes the vocabulary.
% Knowledge distillation from the VideoLLaMA2's visual encoder to the navigation agent's visual encoder was also conducted.
% 
Zhou \etal \cite{Zhou2024NavGPT2UN} pointed out that navigation methods that use foundation models in zero-shot cannot beat learning-based methods, while methods that fine-tune foundation models have the problem of reduced sentence generation ability due to catastrophic forgetting.
Our method further improves the performance of learning-based methods and does not cause catastrophic forgetting. Furthermore, our method is lightweight compared to foundation models, making it capable of running in real time.

Additionally, we experimented with leveraging VideoLLaMA2 not only in a zero-shot setting but also by fine-tuning it on the R2R dataset \cite{anderson2018vision,krantz2020beyond} using QLoRA \cite{dettmers2024qlora}. The number of epochs was set to 1, the learning rate to $2.0\times 10^{-5}$, LoRA $r$ to $128$, and LoRA $\alpha$ to $256$.
See the supplementary materials for more details.

%% file: sec/5_experiment.tex
\section{Experiments}
\label{sec:experiments}

\vspace{-1.5mm}
\subsection{Implementation Details}
\vspace{-1.5mm}
% \kon{スペース的に厳しそうならappendixに持っていく。この論文と同じ設定とだけ書いて、あとはsupple参照されたいとか書く。}

\noindent \textbf{Object-goal Navigation:}
%%\noindent \textbf{Simulation: }
Habitat Simulator \cite{savva2019habitat} and Matterport3D \cite{chang2017matterport3d} scene dataset were used to train RL agents.
%%Habitat Simulator is a simulator commonly used in object-goal navigation.
%%Matteport3D is also a scene dataset commonly used in this task.
%%It is an indoor environment dataset with a 
The dataset consists of large floors with an average of $512\ \rm{m}^2$.
We divide the 67 scenes into 56/4/7 splits for train/val/test, respectively.
%
% Reward is defined as per the default settings of habitat \cite{savva2019habitat}. That is, $2.5$ is given for successfully reaching the goal, and the amount of decrease in the geodesic distance to the goal. In addition, to encourage more efficient goal-reaching, a penalty of $-0.001$ is given at every step.
If $\mathbb{I}_\mathrm{goal}$ represents whether or not the agent has reached the goal, $d_t$ represents the geodesic distance from the agent to the goal at time $t$, and $r_\mathrm{penalty}$ represents the time penalty, then the reward at time $t$ can be expressed as $r_t = \alpha \mathbb{I}_\mathrm{goal} + (d_t - d_{t-1}) + r_\mathrm{penalty}$. Here, the coefficient $\alpha$ 
 and the time penalty are set to $2.5, -0.001$.
%
%%\noindent \textbf{Metrics: }
We used three commonly used metrics: Success Rate (SR), Success rate weighted by Path Length (SPL), Distance To Goal (DTG).
SR represents the rate of the agent reaching goal.
SPL is the standard navigation metric \cite{anderson2018evaluation}, which is the success rate weighted by the ratio of the shortest path to the actual agent path length.
In other words, success with a route that is close to the shortest path results in a higher SPL.
DTG is the distance from the agent location at the end of the episode to the goal.

\noindent \textbf{Vision-and-language Navigation:}
Matterport3DSimulator \cite{anderson2018vision} was used to train IL agents. The entire setup is exactly the same as in the previous study \cite{chen2022think,wang2023scaling}.
%
%%\noindent \textbf{Metrics: }
The same evaluation metrics are used as in the previous study \cite{wang2023scaling}; Navigation Error (NE), SR, and SPL. Here, NE is same as DTG.

\noindent \textbf{Semantic Audio-visual Navigation:}
Soundspaces \cite{chen2020soundspaces} simulator and Matterport3D \cite{chang2017matterport3d} scene dataset were used to train RL agents.
%%Soundspaces is a simulator commonly used in audio-visual navigation, which is an extension of the visual rendering simulator Habitat Simulator \cite{savva2019habitat}.
%%Matteport3D is also a scene dataset commonly used in this task.
We divide the 102 scenes into 73/11/18 splits for train/val/test, respectively.
% It is an indoor environment dataset with a large floor with an average of $512\ \rm{m}^2$.
% There are 73 of these scene datasets available for training, 11 for evaluation, and 18 for testing, each using different scene data.
The setup is exactly the same as in the previous study \cite{chen2021semantic}.
%
% Reward is defined in the same way as in the previous study \cite{chen2021semantic}.
% That is, $+10$ is given for successfully reaching the goal, and $+1$ for reducing the geodesic distance to the goal.
% To encourage more efficient goal-reaching, a penalty of $-0.01$ is also given at every step.
Using the indicator function $\mathbb{I}[\cdot]$, the reward at time $t$ can be expressed as $r_t = \alpha \mathbb{I}_\mathrm{goal} + \mathbb{I}
[d_t > d_{t-1}] + r_\mathrm{penalty}$. Here, $\alpha$ and $r_\mathrm{penalty}$ are set to $10, -0.01$, respectively.
%
%%\noindent \textbf{Metrics:}
The same evaluation metrics are used as in the previous study \cite{chen2020soundspaces}.
That is, SR, SPL, Success rate weighted by Number of Actions (SNA), DTG, and Success When Silent (SWS).
SNA is the success rate weighted by the ratio of the minimum number of actions and the number of agent actions.
% In other words, SNA is higher when the agent succeeds in the number of actions that are close to the minimum number of actions.
SWS is the success rate when the sound stops.
% SR, SPL, SNA, and SWS all take values between 0 and 1, with higher values indicating better performance.
% DTG takes a value greater than or equal to 0. The lower the value, the better the performance.
In this study, the average value over 1,000 tests was obtained.

% \vspace{-1.5mm}
% \subsection{Baselines}
% \vspace{-1.5mm}

% \vspace{-1.5mm}
\subsection{Navigation Performance}
\vspace{-1.5mm}
%
% \vspace{-1.5mm}
\subsubsection{Object-goal Navigation}
\vspace{-2.3mm}
% \noindent \textbf{Does action description prediction contribute to ObjNav?}
\label{sec:objnav}

Two baselines were used.
\noindent \textbf{1) GRU-based \cite{savva2019habitat}} is the simplest method for this task. 
It is based on GRU \cite{chung2014empirical} and is trained with end-to-end RL.
\noindent \textbf{2) SMT \cite{fang2019scene}} has a transformer-based structure and learns by RL.
By keeping visual information in a memory, this considers history.

We compared the proposed method Past-DescRL with the baseline methods.
The proposed method was applied to SMT \cite{fang2019scene}.
As shown in Table \ref{tab:objnav}, our method improved baseline performance on all metrics. 
%%So we showed that explanation prediction contributes to ObjNav.

\begin{table}[tb]
    \centering
    \caption{
        Comparison on object-goal navigation.
        PT indicates whether or not the ADPredictor pre-training (step 1 of phase 2) has been done. $N_\mathrm{SD}$ indicates the number of sharing decoders.
    }
    \vspace{-2mm}
    \label{tab:objnav}
    \scriptsize
    \scalebox{1.05}{
    \begin{tabular}{@{}cccccccc@{}}
        \toprule
        & Method & PT & $N_\mathrm{SD}$ & SR $\uparrow$ & SPL $\uparrow$ & DTG $\downarrow$  \\
        \midrule
        1) & GRU-based \cite{savva2019habitat} & - & - & 7.4 & 4.2 & 6.49 \\
        2) & SMT \cite{fang2019scene} & - & - & 17.9 & 7.7 & 6.72 \\
        3) & SMT w/ Past-DescRL & $\times$ & 0 & 12.8 & 5.7 & 6.95 \\
        4) & SMT w/ Past-DescRL & $\checkmark$ & 0 & \textbf{26.7} & \textbf{9.7} &\textbf{5.91} \\
        5) & SMT w/ Past-DescRL & $\checkmark$ & 2 & 22.8 & 8.0 & \textbf{5.91} \\
        \bottomrule
    \end{tabular}
    }
\end{table}

% \subsubsection{Vision-and-language Navigation}
% \noindent \textbf{Does action description prediction contribute to VLN?}
\vspace{-2.5mm}
\subsubsection{Vision-and-language Navigation}
\vspace{-1.5mm}
\label{sec:vln}
\begin{table}[tb]
    \setlength{\tabcolsep}{1.5pt}
    \centering
    \caption{
        Comparison on vision-and-langauge navigation.
    }
    \vspace{-2mm}
    \label{tab:vln}
    \scriptsize
    \scalebox{1.05}{
    \begin{tabular}{@{}cccccccccc@{}}
        \toprule
        & \multicolumn{3}{c}{Val Seen} & \multicolumn{3}{c}{Val Unseen} &\multicolumn{3}{c}{Test Unseen} \\
        \cmidrule(lr){2-4} \cmidrule(lr){5-7} \cmidrule(lr){8-10}
        Method & NE $\downarrow$ & SR $\uparrow$ & SPL $\uparrow$ & NE$\downarrow$ & SR $\uparrow$ & SPL $\uparrow$ & NE $\downarrow$ & SR$\uparrow$ & SPL $\uparrow$ \\
        \midrule
        Human & - & - & - & - & - & - & 1.61 & 86 & 76 \\
        Random & 9.49 & 16.27 & 14.91 & 9.22 & 16.01 & 14.01 & - & - & - \\
        Seq2Seq \cite{anderson2018vision} & 5.87 & 37.45 & 32.36 & 8.01 & 21.24 & 18.00 & - & -& - \\
        \midrule
        DUET \cite{chen2022think} & 2.26 & 79.73 & 74.78 & 3.21 & 71.65 & 60.44 & 3.63 & 69.76 & 59.39 \\
        w/ Future-DescRL & 2.43 & 80.02 & 74.03 & 3.20 & 71.73 & 60.04 & - & - & - \\
        w/ Past-DescRL & 2.12 & 81.00 & 76.27 & 3.09 & 72.33 & 61.37 & 3.56 & 69.88 & 59.40 \\
        \midrule
        ScaleVLN \cite{wang2023scaling} & 1.95 & 82.57 & \textbf{77.09} & 2.40 & 78.63 & 69.15 & \textbf{2.61} & 77.14 & \textbf{67.34} \\
        w/ Future-DescRL & 2.02 & 81.29 & 75.78 & 2.44 & 78.46 & \textbf{69.48} & - & - & - \\
        w/ Past-DescRL & \textbf{1.88} & \textbf{82.66} & 76.46 & \textbf{2.37} & \textbf{78.84} & 68.96 & 2.64 & \textbf{77.47} & 66.96 \\
        \bottomrule
    \end{tabular}
    }
\end{table}

Four baselines were used.
\noindent \textbf{1) Random} turns to a randomly selected heading and then moves forward. In total, it takes five actions.
\noindent \textbf{2) Seq2Seq \cite{anderson2018vision}} is the simplest model with an encoder-decoder structure. It outputs actions by inputting an instruction into the encoder and visual observations into the decoder.
\noindent \textbf{3) DUET \cite{chen2022think}} has a dual-scale graph transformer for simultaneous long-term action planning and detailed cross-modal understanding. It builds a topological map in real-time to enable efficient exploration in the global action space.
\noindent \textbf{4) ScaleVLN \cite{wang2023scaling}} is a data augmentation method using HM3D \cite{ramakrishnan2021habitatmatterport} and Gibson \cite{xia2018gibson}, scene datasets that are different from Matteport3D used in R2R. This achieved state-of-the-art by combining with DUET.

We compared the proposed method Past-DescRL and Future-DescRL with the baseline methods.
The proposed method was applied to DUET \cite{chen2022think} and ScaleVLN \cite{wang2023scaling}.
As shown in Table \ref{tab:vln}, Past-DescRL slightly improves performance on all metrics for DUET, and the performance improvement is even smaller for ScaleVLN. Since the proposed method is an auxiliary method, its contribution to the main task may be considered small if the task is simple or if the method already performs well. 
While it is true that the SPL of ScaleVLN is not high enough, we believe the difficulty of navigation tasks cannot be fully captured by SPL alone. It also involves the complexity of action decision-making from the robot's perspective. In this regard, SAVNav presents a particular challenge, as the sound stops midway, making it difficult to determine appropriate actions in the latter part of the task. In contrast, tasks such as ObjNav and VLN consistently provide a target object or a language instruction, which facilitates easier decision-making. Consequently, we observe a significant performance improvement in SAVNav, where both the baseline performance and decision-making ease are lower than the other tasks.
On the other hand, it is noteworthy that the proposed method does not reduce the performance of the main task, even though it adds the ability to generate action descriptions.

% \subsubsection{Semantic Audio-visual Navigation}
% \noindent \textbf{Does action description prediction contribute to SAVNav?}
\vspace{-2.5mm}
\subsubsection{Semantic Audio-visual Navigation}
\vspace{-2mm}
\label{sec:savnav}
Five baselines were used.
\noindent \textbf{1) Random}
randomly selects an action from $\{MoveForward, TurnLeft, TurnRight \}$ and $Stop$ if it reaches within a $1\ \rm{m}$ radius of the goal.
\noindent \textbf{2) AV-Nav \cite{chen2020soundspaces}}
is the simplest method first proposed for audio-visual navigation.
It is based on GRU and is trained with end-to-end RL.
\noindent \textbf{3) AV-WaN \cite{chen2020learning}}
learns to generate waypoints using map information.
This one is also based on GRU and is learned by end-to-end RL.
\noindent \textbf{4) SAVi \cite{chen2021semantic}}
is the first method for semantic audio-visual navigation.
It has a transformer-based structure and learns by RL.
It also has a module called Goal Descriptor Network, which predicts goal position and goal category from auditory observations.
\noindent \textbf{5) KSAVEN \cite{tatiya2022knowledge}}
is a knowledge-driven method that uses prior knowledge by utilizing a pre-created knowledge graph.
It has a transformer-based structure and learns by RL.
The knowledge graph is processed by using a graph convolution network \cite{kipf2017semisupervised}. This achieved state-of-the-art performance. The original paper \cite{tatiya2022knowledge} states that all sounds (including unheard) are used to pre-train the audio network, but in this study, only heard sounds were used for fairness with other baselines and making the setting realistic.
% Also, the original paper used its own episode dataset, but for fairness, we also used SAVNav's \cite{chen2021semantic} original episode dataset.

\begin{table}
    \setlength{\tabcolsep}{0.5pt}
    \centering
    \caption{
        Comparison of state-of-the-art methods and DescRL.
        In the Heard/Unheard settings, we tested using sounds heard/not heard during training, respectively. 
    }
    \vspace{-2mm}
    \label{tab:savi}
    \scriptsize
    \scalebox{1.05}{
    \begin{tabular}{@{}ccccccccccc@{}}
        \toprule
        & \multicolumn{5}{c}{Heard} & \multicolumn{5}{c}{Unheard} \\
        \cmidrule(lr){2-6} \cmidrule(lr){7-11}
        Method & SR$\uparrow$ & SPL$\uparrow$ & SNA$\uparrow$ & DTG$\downarrow$ & SWS$\uparrow$ & SR$\uparrow$ & SPL$\uparrow$ & SNA$\uparrow$ & DTG$\downarrow$ & SWS$\uparrow$ \\ \midrule
        Random & 6.4 & 1.8 & 0.7 & 16.5 & 6.2 & 6.4 & 1.8 & 0.7 & 16.5 & 6.2 \\
        AV-Nav \cite{chen2020soundspaces} & 19.3 & 15.9 & 15.0 & 12.6 & 5.6 & 15.7 & 12.8 & 11.9 & 12.6 & 4.7 \\
        AV-WaN \cite{chen2020learning} & 15.9 & 12.3 & 11.6 & 11.0 & 6.1 & 12.5 & 9.9 & 9.2 & 11.2 & 5.1 \\
        \midrule
        SAVi \cite{chen2021semantic} & 31.6 & 28.5 & 24.6 & 11.8 & 12.5 & 24.7 & 22.4 & 18.9 & 11.8 & 10.2 \\
        w/ F-DescRL & 36.4 & 30.8 & 26.6 & \textbf{8.4} & 17.7 & 22.5 & 19.1 & 16.8 & 9.0 & 9.8 \\
        w/ PF-DescRL & 32.6 & 27.3 & 22.2 & 9.7 & 17.0 & 29.4 & 24.4 & 19.5 & 10.2 & \textbf{15.2} \\
        w/ P-DescRL & \textbf{37.4} & \textbf{32.4} & \textbf{28.0} & \textbf{8.4} & \textbf{19.1} & \textbf{31.4} & \textbf{26.9} & \textbf{22.5} & \textbf{8.7} & 15.1 \\
        \midrule
        KSAVEN \cite{tatiya2022knowledge} & 25.1 & 18.1 & 13.5 & 10.3 & 15.8 & 21.1 & 14.9 & 11.5 & 11.2 & 14.2 \\
        w/ F-DescRL & 27.4 & 20.4 & 17.2 & 12.2 & 14.4 & 20.6 & 14.9 & 12.6 & 12.8 & 10.6 \\
        w/ P-DescRL & 21.4 & 15.4 & 11.2 & 12.4 & 13.8 & 21.2 & 15.3 & 10.9 & 12.0 & 14.5 \\
        \bottomrule
    \end{tabular}
    }
\end{table}

We compared the proposed methods P-DescRL, F-DescRL and PF-DescRL with the baseline methods.
The proposed method was applied to SAVi \cite{chen2021semantic}
%, the simplest off-the-shelf method of semantic audio-visual navigation,
and KSAVEN \cite{tatiya2022knowledge}.
%, the current state-of-the-art method.
As shown in Table \ref{tab:savi},
our method achieved state-of-the-art on all evaluation metrics.
%%Here, since the baseline does not use the R2R dataset \cite{anderson2018vision,krantz2020beyond}, our method has more data available for training than the baseline. However, since this is only used to train for generating explanations, not navigation, it is not obvious that this will improve navigation performance.
%%This result indicates that navigation and explainability are highly related and compatible.
%
We also found that learning to generate P-AD has a better impact on navigation than learning to generate F-AD. We believe this is because F-AD prediction is a difficult task in itself and therefore unsuitable as an auxiliary task. We also find that F-DescRL performs worse than the baseline in unheard setting. F-DescRL performs better when it is easy to infer what to do in the future, but not when it is hard to infer what to do in the future. Thus, in the unheard setting, where it is difficult to predict goal categories from sound and hard to infer what to do in the future, performance is expected to be worse than in the baseline. Conversely, F-DescRL outperforms PF-DescRL in heard setting because goal categories are easier to infer and it is easier to infer what to do in the future.

\noindent \textbf{Qualitative results}
% \begin{figure*}[t]
%     \begin{center}
%         \centering
%         \includegraphics[width=\linewidth]{fig/navigation_qualitative_results.pdf}
%     \vspace{-2em}
%         \caption{
%             Navigation trajectories. 
%             Comparing the left and middle figures shows that DescRL reduces wasteful actions and improves performance.
%             The bottom right figure shows a failure case where the agent got close to the goal but failed to stop at the exact position.
%             \kon{TODO: 軌道比較。ここかappendixか。}
%         }
%         \label{fig:navigation_qualitative_result}
%     \end{center}
% \end{figure*}
\begin{figure*}[t]
  \centering
  \begin{minipage}[c]{0.58\textwidth}
    \begin{center}
        \centering
        \includegraphics[width=\linewidth]{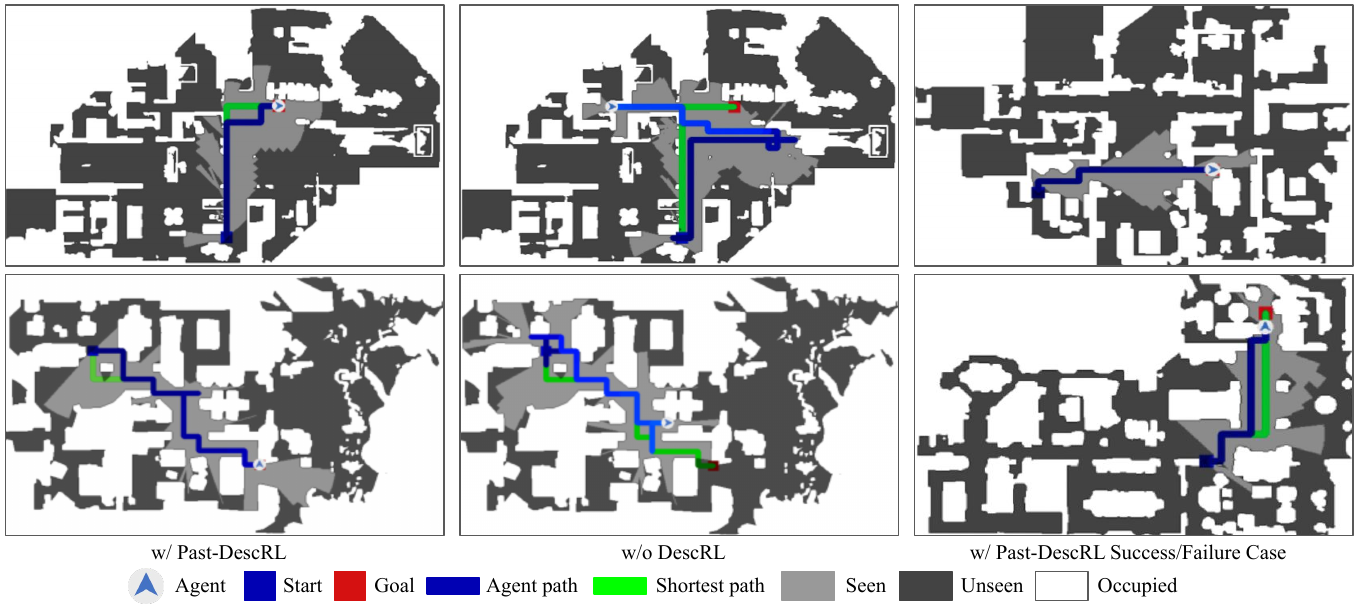}
        \vspace{-6mm}
        \caption{
            Navigation trajectories. 
            Comparing the left and middle figures shows that DescRL reduces wasteful actions.
            The bottom right figure shows a failure case where the agent got close to the goal but failed to stop at the exact position.
        }
        \label{fig:navigation_qualitative_result}
    \end{center}
  \end{minipage}
  \hfill
  \begin{minipage}[c]{0.39\textwidth}
    \begin{center}
        \centering
        \includegraphics[width=\linewidth]{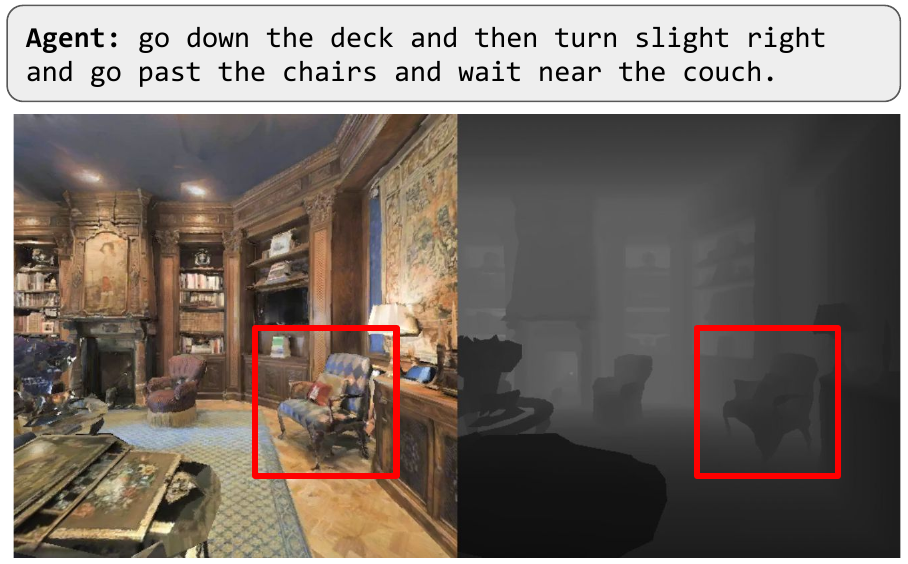}
        \vspace{-6mm}
        \caption{
            Qualitative evaluation of action descriptions. The agent generates the above action descriptions when it observes the RGBD image.
        }
        \label{fig:explainability_qualitative_result}
    \end{center}
  \end{minipage}
  \vspace{-2mm}
\end{figure*}
Figure \ref{fig:navigation_qualitative_result} shows the results of the qualitative evaluation for navigation of SAVi~\cite{chen2021semantic} w/ P-DescRL.
The evaluation was based on unseen scenes and unheard sounds during training.
This result shows that the P-DescRL reduces wasteful actions and enables more accurate navigation.
We believe this is because the action description prediction task allows the agent to learn more important knowledge for navigation.
The bottom right of Fig. \ref{fig:navigation_qualitative_result} shows a case of P-DescRL failure.
In P-DescRL, there were often failure cases where the agent went close but could not stop at the exact position. 
We believe this is due to the domain gap between VLN and SAVNav. R2R used to train ADGenerator is the dataset for VLN. Here, the distance to the goal considered successful in VLN is 3 m, whereas the distance considered successful in SAVNav is 1 m.
In subsequent experiments, SAVNav is used unless otherwise noted.

% \noindent \textbf{Is it superior to other auxiliary tasks?}
\vspace{-1.5mm}
\subsubsection{Comparison to other auxiliary tasks}
\vspace{-1.5mm}
\label{sec:other_aux}
We investigated whether action description prediction is a particularly effective auxiliary task compared to other alternatives. For comparison, we used a supervised learning approach to learn to predict the next action, the progress \cite{ma2019selfmonitoring}, the next image observation, the next auditory observation, the goal location, and the goal category. The results are shown in Table \ref{tab:other_aux_task}. Since the proposed method outperforms the others in most of the metrics, it is clear that the proposed method is a particularly good auxiliary task. 
A semantic understanding of the environment is important for navigation, and we believe that the proposed method promotes this. 
The proposed method is highly beneficial in terms of not only improving navigation performance but also outputting natural language that can be interpreted by humans.

\begin{table}
    \setlength{\tabcolsep}{0.5pt}
    \centering
    \caption{
        Comparison with other auxiliary tasks.
    }
    \vspace{-2mm}
    \label{tab:other_aux_task}
    \scriptsize
    \begin{tabular}{@{}ccccccccccc@{}}
        \toprule
        & \multicolumn{5}{c}{Heard} & \multicolumn{5}{c}{Unheard} \\
        \cmidrule(lr){2-6} \cmidrule(lr){7-11}
        Method & SR$\uparrow$ & SPL$\uparrow$ & SNA$\uparrow$ & DTG$\downarrow$ & SWS$\uparrow$ & SR$\uparrow$ & SPL$\uparrow$ & SNA$\uparrow$ & DTG$\downarrow$ & SWS$\uparrow$ \\ \midrule
        SAVi \cite{chen2021semantic} & 31.6 & 28.5 & 24.6 & 11.8 & 12.5 & 24.7 & 22.4 & 18.9 & 11.8 & 10.2 \\
        w/ Next Action      & 33.2 & 30.6 & 27.3 & 10.7 & 12.7 & 23.1 & 20.8 & 17.8 & 11.1 & 9.3 \\
        w/ Progress \cite{ma2019selfmonitoring}        & 35.0 & 31.6 & \textbf{28.3} & 10.2 & 15.7 & 24.2 & 21.1 & 18.3 & 11.0 & 9.4 \\
        w/ Next Frame       & 35.4 & 31.6 & 27.0 & 11.0 & 15.8 & 24.7 & 21.8 & 18.4 & 11.5 & 9.9 \\
        w/ Next Spectrogram & 34.7 & 31.3 & 27.3 & 11.4 & 13.4 & 23.9 & 21.0 & 18.1 & 12.0 & 8.8 \\
        w/ Goal Location    & 34.5 & 30.2 & 25.8 & 10.3 & 15.6 & 25.7 & 22.3 & 18.5 & 10.7 & 11.2 \\
        w/ Goal Category    & 35.4 & 31.9 & 27.9 & 10.2 & 15.2 & 25.3 & 22.4 & 19.4 & 10.8 & 10.5 \\
        w/ F-DescRL      & 36.4 & 30.8 & 26.6 & \textbf{8.4} & 17.7 & 22.5 & 19.1 & 16.8 & 9.0 & 9.8 \\
        w/ PF-DescRL & 32.6 & 27.3 & 22.2 & 9.7 & 17.0 & 29.4 & 24.4 & 19.5 & 10.2 & \textbf{15.2} \\
        w/ P-DescRL        & \textbf{37.4} & \textbf{32.4} & 28.0 & \textbf{8.4} &\textbf{19.1} & \textbf{31.4} & \textbf{26.9} & \textbf{22.5} & \textbf{8.7} & 15.1 \\
        \bottomrule
    \end{tabular}
    %}
\end{table}

% \noindent \textbf{Can using a foundation model as XGenerator also improve performance without relying on high-cost human data?}
\vspace{-1.5mm}
\subsubsection{VLM as ADGenerator}
\vspace{-1.5mm}
\label{sec:vlms}
The results are shown in Table \ref{tab:foundation_model}. The best performance was found when using the ADGenerator, which has a simple structure learned from scratch using R2R. However, even in the case without any human-created data (row 3), we found that DescRL can improve the performance. This indicates that our method is able to implicitly extract the knowledge necessary for navigation from VLMs without any human-created data. We also found that the performance was lower when we used the VLM fine-tuned by R2R dataset. This may be due to the VLM being overfitted to the dataset.
Fothermore, when comparing row 3 and row 5, using  a stronger foundation model (Qwen2.5-VL) does not necessarily lead to better performance.

\begin{table}[tb]
    \setlength{\tabcolsep}{0.5pt}
    \centering
    \caption{
        Comparison of using a model trained from scratch and using a pre-trained VLM as the ADGenerator (ADGen).
        CNN-TF, VL2, and QVL indicate CNN+Transformer, VideoLLaMA2, and Qwen2.5-VL respectively.
        FT indicates whether the model was fine-tuned (or trained from scratch) by the R2R dataset.
    }
    \vspace{-2mm}
    \label{tab:foundation_model}
    \scriptsize
    \scalebox{1.05}{
    \begin{tabular}{@{}ccccccccccccc@{}}
        \toprule
        & & & \multicolumn{5}{c}{Heard} & \multicolumn{5}{c}{Unheard} \\
        \cmidrule(lr){4-8} \cmidrule(lr){9-13}
        & ADGen & FT & SR$\uparrow$ & SPL$\uparrow$ & SNA$\uparrow$ & DTG$\downarrow$ & SWS$\uparrow$ & SR$\uparrow$ & SPL$\uparrow$ & SNA$\uparrow$ & DTG$\downarrow$ & SWS$\uparrow$ \\
        \midrule
        1) & None & - & 31.6 & 28.5 & 24.6 & 11.8 & 12.5 & 24.7 & 22.4 & 18.9 & 11.8 & 10.2 \\
        2) & CNN+TF & $\checkmark$ & \textbf{37.4} & \textbf{32.4} & \textbf{28.0} & \textbf{8.4} & \textbf{19.1} & \textbf{31.4} & \textbf{26.9} & \textbf{22.5} & \textbf{8.7} & \textbf{15.1} \\
        3) & VL2 & $\times$ & 33.7 & 29.8 & 24.4 & 10.6 & 16.0 & 28.9 & 24.3 & 19.5 & 10.6 & 14.2 \\
        4) & VL2 & $\checkmark$ & 28.9 & 25.6 & 21.6 & 11.6 & 11.6 & 23.3 & 20.3 & 16.7 & 11.9 & 10.2 \\
        5) & QVL & $\times$ & 33.4 & 28.6 & 25.6 & 8.9 & 15.2 & 28.2 & 23.8 & 20.5 & 9.3 & 11.5 \\
        \bottomrule
    \end{tabular}
    }
\end{table}

% \noindent \textbf{Is it useful for other navigation tasks?}

% We investigated whether our method is useful not only for audio-visual navigation but also for other navigation tasks. Table \ref{tab:objnav} shows the results when applied to object-goal navigation using only visual observations, and Table \ref{tab:vln} shows the results when applied to vision-and-language navigation. \kon{TODO: results and discussion}

\vspace{-1.5mm}
% \subsection{Action Description Results}
\subsection{Analysis of Failures by Action Description}
\vspace{-1.5mm}

Here, we focus on the action descriptions generated by the ADPredictor in the failed episode and analyze why it failed.
It generates action description based on its own output, the probability of each word $\hat{p}^S (w_l=w|w_1, ..., w_{l-1}), (w \in \mathrm{Vocab})$, using greedy search in VLN and using top-k sampling \cite{fan2018hierarchical} and top-p sampling \cite{Holtzman2020The} where the temperature $\tau = 2.0$ and $k=10, p=0.95$ in SAVNav.

\begin{figure}[t]
    \begin{center}
        \centering
        \includegraphics[width=\linewidth]{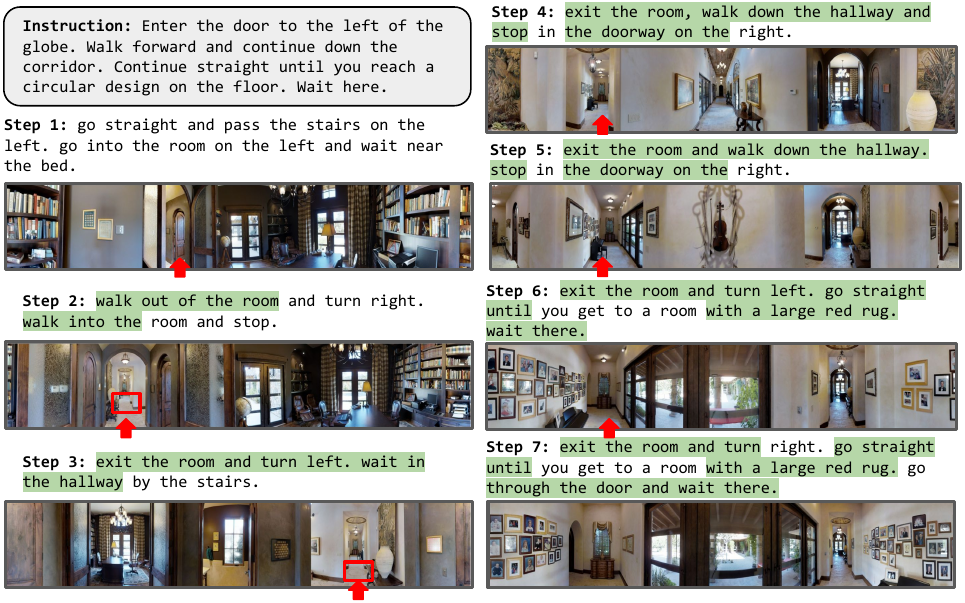}
        \vspace{-2em}
        \caption{
            Qualitative evaluation of generating action descrptions performance of ScaleVLN \cite{wang2023scaling} w/ Past-DescRL on VLN.
            The agent observes the panoramic image at each step and must follow the given instructions. The red arrows represent the actions selected by the agent, and the action descriptions shown above each panorama image are ones generated by the agent. Areas marked in green are action descriptions that can be judged to be correct.
            % In this episode, the agent failed to navigate.
            %The agent had to stop at step 5. The generated action description shows that the agent was not able to focus on the circle on the floor with the red frame, which is specified in the instruction.
        }
        \vspace{-2em}
        \label{fig:vln_x_qualitative_eval_failure}
    \end{center}
\end{figure}

Figure \ref{fig:explainability_qualitative_result} shows the RGBD image observed by the agent and the action desciption during the failure episode in the bottom right in Fig. \ref{fig:navigation_qualitative_result}. The agent outputs ``wait near the couch'' even though it is still far from the goal chair framed in red. It can be analyzed that the agent stopped early because it judged that it was close enough to the chair, even though it was still far. 
%%While the proposed method was able to analyze the failures like that, it was also found that object misrecognition and left-right misrecognition occurred frequently. 

Figure \ref{fig:vln_x_qualitative_eval_failure} shows a result of the qualitative evaluation of action descriptions of ScaleVLN \cite{wang2023scaling} w/ Past-DescRL on VLN. Here, it is evaluated with Val Unseen dataset.
% It shows the results for failed navigation. 
In the episode in Fig. \ref{fig:vln_x_qualitative_eval_failure}, the agent stopped at the 7th step, but actually had to stop at 5th step. In other words, as the instruction says, the agent needed to stop on the floor circle enclosed by the red square, but was unable to do so. When we look at the action descriptions predicted by the agent, we can see that it did not pay attention to the circle on the floor at all. Therefore, it can be analyzed that the agent could not stop at the correct position because it could not pay attention to the circle on the floor.
% Also, although no stairs were observed in the observations, the word ``stairs'' is often found in  action descriptions. This may be because the agent has learned the common  sense that stairs are often found in hallways.

Quantitative evaluation and more detailed qualitative evaluation can be found in the supplementary material.

\vspace{-1.5mm}
\subsection{Ablation Study}
\vspace{-1.5mm}

%%We ablate the core design choices applied in this paper. In particular, we investigated the need for XPredictor pre-training (step 1 of phase 2), the need for task embeddings, and the number of decoders to be shared. The results are shown in Table \ref{tab:ablation}.

\noindent \textbf{ADPredictor Pre-training: }
Comparing rows 1 and 2, rows 4 and 6 in Table \ref{tab:ablation}, and rows 3 and 4 in Table \ref{tab:objnav}, we can see that the performance is always better with pre-training (step 1 of phase 2). Therefore, we can see that this is a very important factor. This is because the learning unrelated to navigation, such as grammar, is done in advance.

\noindent \textbf{Task Embedding: }
Comparing rows 5 and 6 of Table \ref{tab:ablation}, we can see that the performance is higher with task embeddings for most of the metrics. Thus, we can see that this is also an important factor. This is important in determining whether the agent is now trying to decide on an action or predict an action description.

\noindent \textbf{The number of sharing decoders: }
We investigate how many layers of shared decoders would improve SAVNav performance the most when an agent has three layers of decoders. Comparing rows 2, 3, 6, and 7 of Table \ref{tab:ablation}, we found that sharing two out of three layers improves SAVNav performance for many of the metrics. If no decoders are shared at all, the benefit to navigation from action description prediction is reduced.
% This suggests that the performance did not improve when the number of shared layers was small. 
On the other hand, if all decoders are shared, the latent space for decision-making and the latent space for predicting action descriptions must be equivalent. However, this is contrary to the fact. Therefore, conversely, the case where all decoders are shared would also not have resulted in higher accuracy.
This result indicates that a trade-off still exists between navigation and faithfulness
%; however, we believe this gap can be narrowed as the accuracy of sentence generation improves
.
Also, in ObjNav, the performance was higher when the number of shared decoders was smaller (row 4 and row 5 in Table \ref{tab:objnav}). This suggests that SAVNav is a particularly difficult task, and therefore it is especially important to supplement it with other tasks.

% \begin{figure}[t]
%     \begin{center}
%         \centering
%         \includegraphics[width=\linewidth]{fig/x_qualitative_eval.pdf}
%         \vspace{-6mm}
%         \caption{
%             Qualitative evaluation of action descriptions. The agent generates the above action descriptions when it observes the RGBD image.
%         }
%         \label{fig:explainability_qualitative_result}
%     \end{center}
% \end{figure}

\begin{table}
    \setlength{\tabcolsep}{0.5pt}
    \centering
    \caption{
        Results of the ablation experiment.
        PT indicates whether the ADPredictor is pre-trained. TE indicates whether task embedding is used. $N_\mathrm{SD}$ indicates the number of sharing decoders.
    }
    \vspace{-2mm}
    \label{tab:ablation}
    \scriptsize
    \scalebox{1.05}{
    \begin{tabular}{@{}ccccccccccccccc@{}}
        \toprule
        & & & & \multicolumn{5}{c}{Heard} & \multicolumn{5}{c}{Unheard} \\
        \cmidrule(lr){5-9} \cmidrule(lr){10-14}
         & PT & TE & $N_\mathrm{SD}$ & SR$\uparrow$ & SPL$\uparrow$ & SNA$\uparrow$ & DTG$\downarrow$ & SWS$\uparrow$ & SR$\uparrow$ & SPL$\uparrow$ & SNA$\uparrow$ & DTG$\downarrow$ & SWS$\uparrow$ \\
        \midrule
        1) & $\times$ & - & 0 & 33.9 & 30.7 & 26.8 & 10.4 & 13.6 & 25.6 & 22.2 & 19.2 & 10.9 & 10.4 \\
        2) & $\checkmark$ & - & 0 & 37.6 & 32.6 & \textbf{28.0} & \textbf{7.8} & 18.6 & 26.6 & 23.7 & 20.3 & \textbf{8.7} & 11.2 \\
        3) & $\checkmark$ & $\checkmark$ & 1 & \textbf{38.7} & \textbf{33.3} & 27.2 & 8.0 & 21.5 & 30.5 & 26.1 & 20.9 & \textbf{8.7} & 16.2 \\
        4) & $\times$ & $\checkmark$ & 2 & 34.1 & 31.3 & 27.6 & 10.3 & 13.3 & 25.1 & 22.3 & 19.4 & 11.0 & 10.4 \\
        5) & $\checkmark$ & $\times$ & 2 & 37.4 & 31.7 & 25.0 & 8.3 & \textbf{21.6} & 30.0 & 24.9 & 20.0 & \textbf{8.7} & \textbf{16.9} \\
        6) & $\checkmark$ & $\checkmark$ & 2 & 37.4 & 32.4 & \textbf{28.0} & 8.4 & 19.1 & \textbf{31.4} & \textbf{26.9} & \textbf{22.5} & \textbf{8.7} & 15.1 \\
        7) & $\checkmark$ & $\checkmark$ & 3 & 30.2 & 26.6 & 19.8 & 9.9 & 15.9 & 26.5 & 23.0 & 17.1 & 10.3 & 14.1 \\
        \bottomrule
    \end{tabular}
    }
    \vspace{-3mm}
\end{table}

\vspace{-2.0mm}
\subsection{Limitations}
\vspace{-1.5mm}
The limitations of our current proposed method are as follows.
1) Harm caused by biased common sense in the dataset. For example, due to the frequent occurrence of stairs in the hallways in the dataset, the word ``stairs'' appeared even if there are no actual stairs when walking down the hallway.
% This could be improved with a more diverse dataset.
%が、一方でこれは人間でも起きてしまう難しい問題でもある。
2) Stopping in front of an object in the true category but in different instances. Due to the lack of qualifiers about the object in the description, 
the category can be identified but 
the specific appearance cannot be learned. 
In our proposed method using VLMs, we can expect improvement 
% by making it possible to generate ``explanations'' rich in modifiers 
by adjusting the prompt to generate action descriptions rich in modifiers.
3) Adaptation beyond navigation. We are interested in whether the proposed method can also be applied to other RL tasks or to Embodied Question Answering \cite{majumdar2023openeqa}.
4) The proposed method may not work if the 
% characteristics of the robot that is navigating are
robot differs from those assumed when creating the training data for generating action descriptions. 
For example, 
it is questionable whether the ADGenerator trained on the R2R dataset, which was created assuming a roughly human-sized robot, can be applied to learning a navigation model for a smaller robot. 
The view changes with the robot's size, so the correct action description should also adjust accordingly.

%% file: sec/6_conclusion.tex
\vspace{-1.5mm}
\section{Conclusion}
\label{sec:conclusion}
\vspace{-1.5mm}
We proposed DescRL, an approach that enables agents to describe their actions while integrating both IL and RL methods.
Our experiments confirmed that action description prediction tends to enhance navigation performance more effectively than other auxiliary tasks.
Also, by leveraging knowledge distillation from a VLM, our experiments suggested that DescRL can be trained without reliance on expensive human-created data, making it a more adaptable method.
%Through comprehensive experiments, 
We demonstrated its consistent effectiveness for various navigation tasks in improving navigation performance while making the navigation descriptive. Furthermore, our method achieved state-of-the-art performance on SAVNav, a highly challenging navigation task, and suggested the possibility of analyzing failure cases using the predicted action descriptions.

% In future work, we are interested in applying the method to different types of robots. We are interested in how changes in the embodiment of the robot, such as size and shape, will change the explanations generated and their impact on the main task.